\pdfoutput=1
\documentclass[10pt,twocolumn,letterpaper]{article}

\usepackage{cvpr}
\usepackage{times}
\usepackage{epsfig}
\usepackage[pdftex]{}
\usepackage{amsmath}
\usepackage{amssymb}

\usepackage[pagebackref=true,breaklinks=true,letterpaper=true,colorlinks,bookmarks=false]{hyperref}
\usepackage{color}
\newcommand{\figref}[1]{Fig. \ref{#1}}
\newcommand{\figrefh}[1]{Figure \ref{#1}}
\newcommand{\tabref}[1]{Table \ref{#1}}

\cvprfinalcopy % *** Uncomment this line for the final submission

\def\cvprPaperID{****} % *** Enter the CVPR Paper ID here

\ifcvprfinal\fi
\begin{document}

%%%%%%%%% TITLE
%\title{\LaTeX\ Author Guidelines for CVPR Proceedings}
\title{Hand Orientation Estimation in Probability Density Form}

\author{Kazuaki Kondo\\
  Kyoto University\\
  Kyoto, Japan\\
%Yoshida-honmachi, Sakyo, Kyoto, Japan\\
{\tt\small kondo@media.kyoto-u.ac.jp}
\and
Daisuke Deguchi\\
Nagoya University\\
Aichi, Japan\\
{\tt\small ddeguchi@nagoya-u.jp}
\and
Atsushi Shimada\\
Kyushu University\\
Fukuoka, Japan\\
{\tt\small atsushi@ait.kyushu-u.ac.jp}
}

\maketitle
%\thispagestyle{empty}

%%%%%%%%% ABSTRACT
\begin{abstract}
  Hand orientation is an essential feature required to understand hand behaviors and subsequently support human activities.
  In this paper, we present a new method for estimating hand orientation in probability density form.
  It can solve the cyclicity problem in direct angular representation and enables the integration of multiple predictions based on different features.
  We validated the performance of the proposed method and an integration example using our dataset, which captured cooperative group work.
\end{abstract}

\renewcommand{\thefootnote}{\fnsymbol{footnote}}
\footnote[0]{Accepted for presentation at The fourth international workshop on Egocentric Perception, Interaction and Computing : EPIC @CVPR2019}
\renewcommand{\thefootnote}{\arabic{footnote}}

%%%%%%%%% BODY TEXT
%--------------------------------------------------------------
\section{Introduction}
Recognizing hands in videos is one of fundamental approaches to understanding human-to-human and human-to-environment interaction.
Previous studies have approached detecting hands in various scenes, including daily snapshots\cite{Mittal2011}, driving vehicles\cite{Le2017} and interaction captured from ego-centric cameras\cite{Bambach2015}..
Currently, the performance of hand detection has almost reached the level of practical use.
For the next step, in this study, we propose an approach to estimate hand orientation in an image as an important attribute for deeper analysis. Hand orientation can provide a better understanding of hand poses or interaction direction.

A major issue when estimating the orientation of an object in an image is the cyclicity in angular representation.
In particular, $0$ and $2\pi$ radians are quite distant in terms of angular representation, but they correspond to the same orientation.
An orientation estimation method that does not consider this cyclicity results in inaccurate angles. 
The experiments presented by Deng et al. \cite{Deng2017} illustrated that direct angle representation leads to poor performance in hand orientation estimation.
Qu et al. \cite{Qu2018} assumed that regression errors are large at a singular angle, but do not propagate to the angles far from it. They used a three-path-way convolutional neural network (CNN) that independently estimates three angular values that correspond to the same orientation in the image, but have different origin angles.
The final output is determined by majority voting of the three estimated angles because one of them may have a great influence on the cyclicity, whereas the remaining two do not. 
Instead of angular representation, Schuch et al. \cite{Schuch2017} and Yang et al. \cite{Yang2018} proposed representing orientation using a vector whose components are projections onto the horizontal and vertical axes. These component values are continuous at an arbitrary angle and their combination has a one-to-one correspondence to the orientation. Thus, cyclicity does not have to be considered. 

In the present study, we propose another representation: probability density form $p(\theta)$, which is different from direct angle $\theta$ or vector $(cos\theta,sin\theta)$ mentioned above.
This representation has the following two advantages:
%(1) Estimating $p(\theta)$ instead of $\theta$ can avert the cyclicity problem. The change of estimation target (dimension) is a similar approach to vector representation.
%(2) In most cases in previous works, only a rectangular hand region was used for estimating hand orientation. However, we often have other valuable features, such as the hand position in the image or relative geometry to other body parts. When multiple estimation results that use individual features are all represented in probability density form, they can be integrated easily. 
\begin{itemize}
\item Estimating $p(\theta)$ instead of $\theta$ can avert the cyclicity problem. The change of estimation target (dimension) is a similar approach to vector representation. 
\item In most cases in previous works, only a rectangular hand region was used for estimating hand orientation. However, we often have other valuable features, such as the hand position in the image or relative geometry to other body parts. When multiple estimation results that use individual features are all represented in probability density form, they can be integrated easily. 
\end{itemize}
\noindent
In this paper, we explain an approach to apply the probability density form to a CNN framework and an example of integrating multiple estimation results. We evaluate the performance of the proposed method on an original dataset that captures cooperative group work.

%--------------------------------------------------------------
\section{Methodology}
\subsection{Orientation estimation in density form}
\begin{figure}[t]
  \centering
  \includegraphics[width=0.90\linewidth]{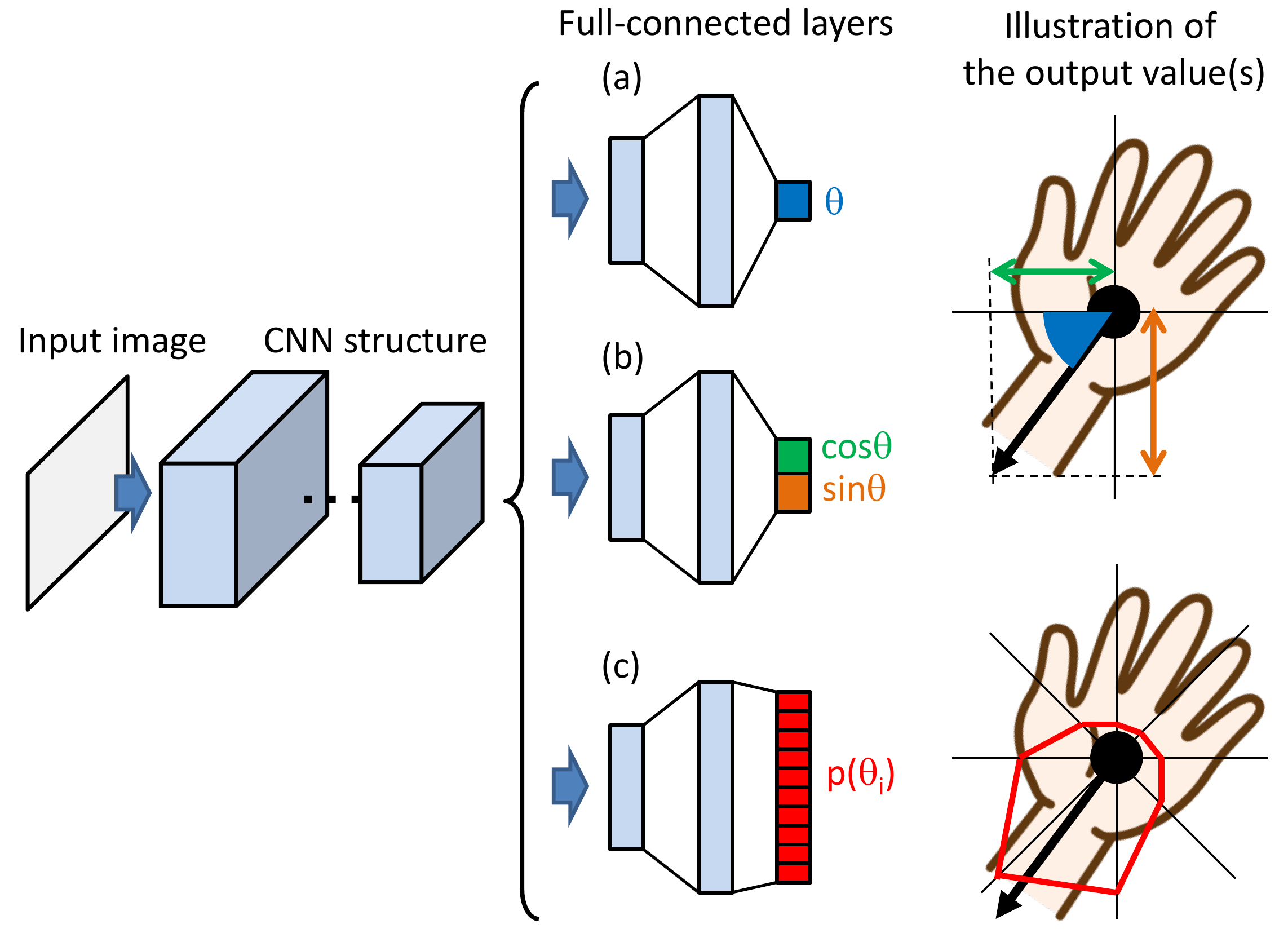}
  \caption{Comparison of CNN structures for hand orientation estimation: (a) direct angular representation; (b) vector presentation; and (c) our probability density form and its illustration in radar chart style.}
  \label{fig:CNN}
\end{figure}

In recent years, CNN-based methods have achieved promising results in generic image recognition.
Thus, we apply our density form approach to a typical image-based CNN that accepts a rectangular image as its input.
The proposed method describes the probability density of hand orientation in a non-parametric and discrete form, $p(\theta_i)$, where $\theta_i = \frac{2 \pi i}{N}, i=0,1,2 ... N-1$.
For this representation, an $N$-path CNN estimates $N$ probability values $p(\theta_i)$ at its output layer instead of one angular value for the direct representation or two projected values for the vector representation (\figref{fig:CNN}).
The softmax activation function is applied at the output layer to meet the probability density condition $\sum_i p(\theta_i) = 1$.
These are the only requirements of the proposed method; we do not have any other constraints on the CNN structure, including its former layers. Our probability density form can be applied to any type of CNN.

Alternatively, we need to convert ground truth orientations into density forms to use them for training the CNN, and recover the orientation angle from the estimated probability density. This mutual conversion is used as follows:

\vspace{2mm}
\noindent
{\bf Orientation angle to probability density}\\
\noindent
Assume not only ground truth orientation $\theta_{gt}$ but also its neighborhood orientation $\theta_{neighbor}$ has a likelihood in $p(\theta_{gt})>p(\theta_{neighbor})>0$ relations; that is, orientations close to the ground truth are accpetable as estimated values.
Using this assumption, probability density $p(\theta_i)$ is determined as
\begin{equation}
\begin{array}{c}
p(\theta_i) = \frac{1}{s_{vd}} exp \left ( -\frac{d(\theta_i,\theta_{gt})^2}{2\sigma^2}\right ) \\
d(\theta_1,\theta_2) = acos( ( cos\theta_1, sin\theta_1 )\cdot(cos\theta_2, sin\theta_2))
\end{array}
\end{equation}
\noindent
from ground truth orientation $\theta_{gt}$, where $\sigma$ denotes the acceptable degree described by the Gaussian distribution. $s_{vd}$ is a normalization term for $\sum_i p(\theta_i)=1$. Through the vector inner product form, angular distance $d(\theta_1,\theta_2)$ averts the cyclicity problem.

%$\sigma$ works like a regularizer in training. Large $\sigma$ provides a meaningful likelihood to the large number of $\theta_i$, which reduces overfitting caused by a small amount of training data.

\vspace{2mm}
\noindent
{\bf Probability density to orientation angle}\\
\noindent
Optimal orientation angle $\theta$ is estimated using an inverse approach to the above process:
\begin{equation}
\theta = argmax_{\theta} \frac{\sum_i p(\theta_i)exp\left( -\frac{d(\theta_i,\theta)}{2\sigma^2}\right)}{\sum_i p(\theta_i)}
\end{equation}
\noindent
based on cosine similarity between $\{p(\theta_i)\}$ and $\mathcal{N}(\theta,\sigma)$.

\subsection{Integration of multiple densities}
The visual pattern of a hand region strongly reflects hand orientation; however, other features are often useful as well. The integration of estimation results using multiple features improves performance.
As an example, we present two additional estimation methods for hand orientation and integrate them into $p(\theta_i|H_{img})$, which is estimated by the proposed method that uses only hand region image $H_{img}$.

\vspace{2mm}
\noindent
{\bf Human region-based estimation: $p(\theta_i|H_r)$ }\\
\noindent
Because the hand is the endpoint of the human body, human region $H_r$ definitely appears in the neighborhood of the hand in the image. This feature provides a somewhat weak but valuable additional constraint for hand orientation.
Probability density $p(\theta_i|H_r)$ is given by the amount of the human region that lies in each orientation $\theta_i$:
\begin{equation}
  \begin{array}{c}
    p(\theta_i|H_r)	= \frac{1}{s_{h}}\sum_{r=R}^{K_R R} H_r(cx+rcos\theta_i,cy+rsin\theta_i) \\
    H_r(x,y) = \left\{ 
    \begin{array}{ll}
      1 & (x,y) \, \text{is in the human region} \\
      0 & \text{otherwise} \\
    \end{array}
    \right .
  \end{array}
\end{equation}
\noindent
where $(cx,cy), R$ and $s_h$ denote the center coordinate of the hand region, half the size of the hand and a normalization term for $\sum_i p(\theta_i|H_r) =1$, respectively. $K_R$ determines the upper limit of the neighborhood region of the hand.

\vspace{2mm}
\noindent
{\bf Position-based estimation: $p(\theta_i|cx,cy)$}\\
\noindent
The appearance and position of the hand in the image often depends on the target scenario or camera location, which is used to provide some constraint on hand orientation.
The tracking and recognition of a hand presented by Lee et al. \cite{Lee2014} used a hand position constraint in first-person videos that captured interactions in a face-to-face style.  
Using a similar idea, the position of the hand provides a constraint on hand orientation.  
To predict the hand orientation probability at an arbitrary position in the image, we use an interpolation of the probability densities at grid points $(gu_k,gv_k)$ in the neighborhood of target position $(cx,cy)$:
\begin{equation}
  p(\theta_i|cx,cy) = \frac{\sum_k w_b(gu_k,gv_k,cx,cy)p(\theta_i|gu_k,gv_k)}{\sum_k w_b(gu_k,gv_k,cx,cy)},
\end{equation}
\noindent
where $w_b(gu,gv,cx,cy)$ denotes a bilinear weight that represents the contribution amount of grid point $(gu,gv)$ to target position $(cx,cy)$. The probability densities at grid points $p(\theta_i|gu_k,gv_k)$ are trained as weighted means of $M$ ground truth samples $p(\theta_i|x_j,y_j), j=1,2,...M $.

When we assume that the visual pattern of the hand region, human region in the neighborhood of the hand and hand position on the image are independent features, their integrated estimation is given by the simple product of the individual probability densities.% as $p(\theta_i|H_{img}, H_r, cx, cy) = p(\theta_i|H_{img})p(\theta_i|H_r)p(\theta_i|cx,cy)$.
The optimal orientation angle that matches the integrated probability density $p(\theta_i|H_{img}, H_r, cx, cy)$ can be extracted using the same approach explained in  Eq. (2).

%--------------------------------------------------------------
\section{Experimental evaluation}
\subsection{Dataset}
\begin{figure}[t]
  \centering
  \includegraphics[width=0.90\linewidth,bb=0 0 640 385 ]{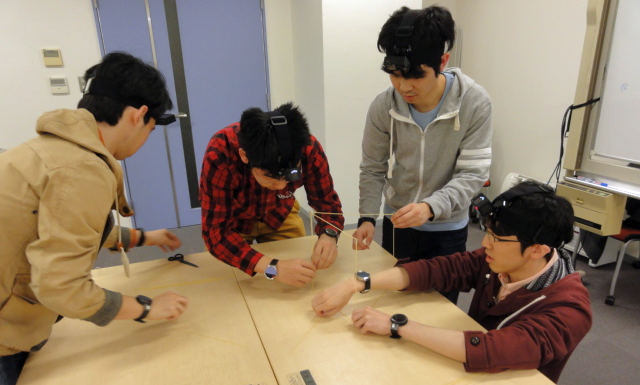}
  \caption{Example scene of the group work task.}
  \label{fig:marshmallow_challenge}
\end{figure}

Our motivation for estimating hand orientation is to use it to analyze interactions in cooperative group work. In the experimental evaluation, we used first-person view videos captured from head-mounted cameras on the group work participants. 
A cooperative assembly task was given to the participants, which was building a tower using dried pastas, adhesive tape and thread, with a marshmallow on the top, known as the ``Marshmallow Challenge'' (\figref{fig:marshmallow_challenge}).
It is often used for ice breaking or community building, but it can be used as a good example to observe cooperative interaction because the participants must build a tower as high as they can with quite limited/weak materials, within short time.
We conducted the group work task for two groups that consisted of four university students. We captured a total of 17 min. $\times$ 8 first-person videos in $1920 \times 1080$ pixels. We manually extracted over 10 thousand hands from the videos with their bounding boxes and orientations running from the palm centers to the wrist centers.
We used this dataset as the ground truth for training and verifying the estimated results.

\subsection{Experimental configuration}
Density form estimation can be applied easily to any type of CNN structure. For the early stage validation, we selected a simple CNN that combined the convolutional part of VGG16\cite{vgg16} and a 1,024-dimension fully connected layer.
The hand region images were resized to $112 \times 112$ pixels to be inserted into the CNN.
The sampling resolution of the orientation and the Gaussian variance for value-density mutual conversion were configured as $N=16$ and $\sigma=10^\circ$, respectively. 
KL divergence was used as the loss function to maximize the distribution similarity during parameter training. 
Training started from the pre-trained parameters using the imageNet dataset. The last layer of the VGG16 convolutional part and the additional fully connected layers were fine-tuned using our dataset. 
To prevent overfitting, we also used the dataset augmentation trick, which applied a combination of random shift and expansion to the original hand images.
In the additional estimation methods, human regions were extracted by the pre-trained PSPNet\cite{pspnet}. The neighborhood regions were defined using $K_R=4.0$, considering the ratio of the hand size and wrist-to-elbow distance.
The grid size for position-based estimation was $5 \times 4$ along the image's horizontal and vertical axes.

The comparative methods were a simply estimating one angle as a baseline, the angle duplication and the vector representation. The last two methods were configured in reference to \cite{Qu2018} and \cite{Yang2018}, respectively.
The mean squared error was used as the loss function in those methods.
%To match the vector representation, the mean squared error was used as the loss function, and L2 normalization was applied to the output layer.
The other conditions were all the same as those used in the proposed method.
Performance validation was conducted using 10-fold cross validation.

\begin{figure*}[t]
  \centering
  \begin{tabular}{ccc}
    \includegraphics[width=0.31\linewidth,bb=0 0 960 540 ]{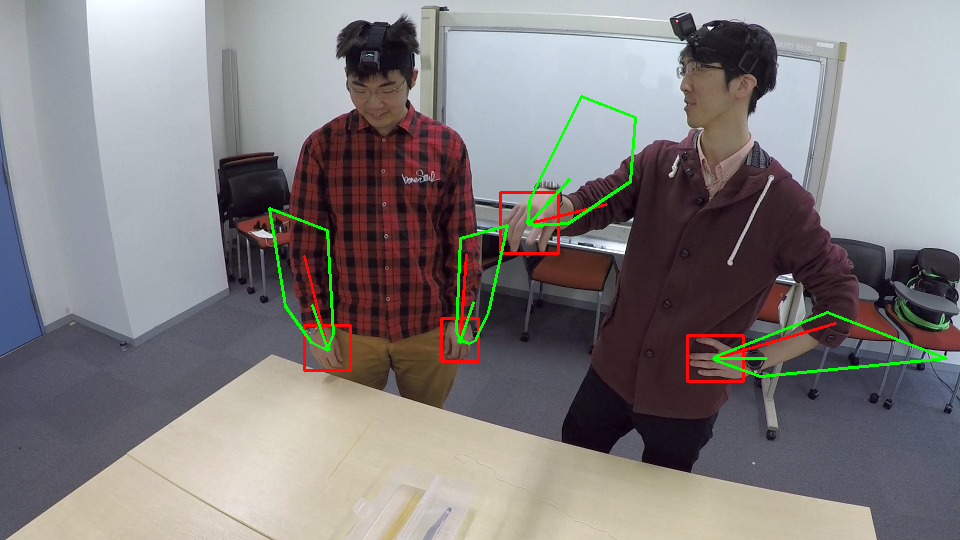} &
    \includegraphics[width=0.31\linewidth,bb=0 0 960 540 ]{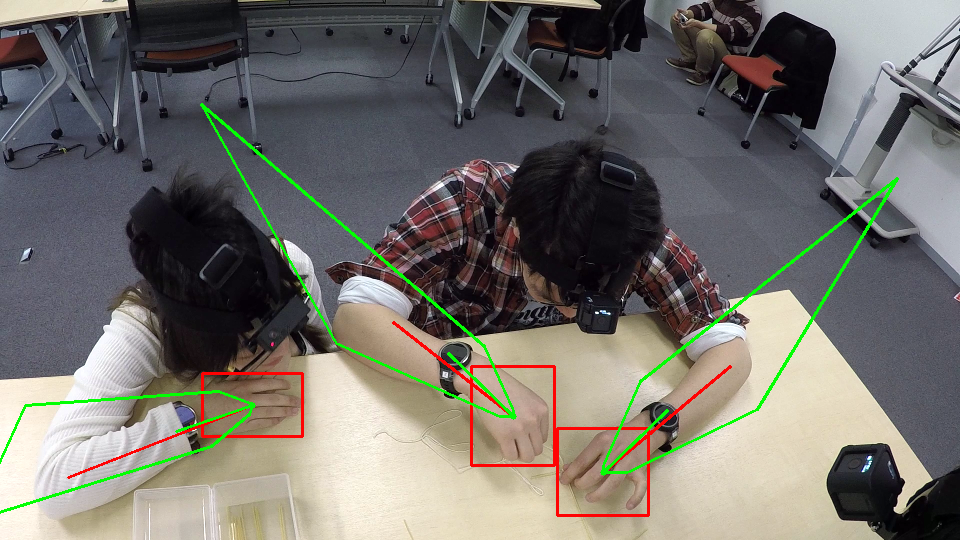} & 
    \includegraphics[width=0.31\linewidth,bb=0 0 960 540 ]{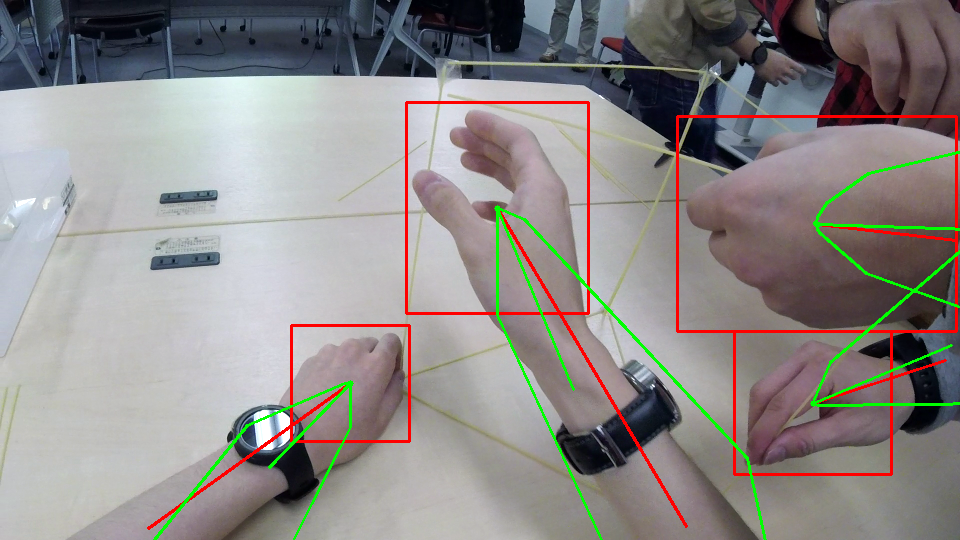} \\
    \end{tabular}
  \caption{Examples of the estimation results. The red lines running from the hand centers denote the ground truth hand orientations. The green charts and lines correspond to estimated probability density $p(\theta_i|H_{img})$ and the final orientation prediction, respectively.}
  \label{fig:correct_results}
\end{figure*}

\begin{table*}[t]
\centering
\caption{Orientation error populations}
\begin{tabular}{|c|c|c|c|c|c|c|c|}
\hline
Method & $<10^\circ$ & $<20^\circ$ & $<30^\circ$ & $\cdots$ & $>90^\circ$ & $>120^\circ$ & $>150^\circ$ \\ 
\hline \hline

baseline & 35.3\% & 61.9\% & 77.5\% &- & 3.22\% & 1.73\% & 0.834\% \\ 
\hline

duplicated angles\cite{Qu2018} & 40.9\% & 69.7\% & 85.2\% &- & 0.702\% & {\bf 0.197\%} & {\bf 0.056\%} \\ 
\hline

vector representation\cite{Yang2018} & 41.3\% & 70.0\% & 85.8\% &- & {\bf 0.448\%} & 0.224\% & 0.126\%\\ 
%vector representation & 49.0\% & 79.0\% & 91.5\% &- & {\bf 0.253\%} & {\bf 0.084\%} & {\bf 0.028\%} \\ 
\hline

density representation & {\bf 61.5\%} & {\bf 88.0\%} & {\bf 95.1\%} &- & 0.731\% & 0.431\% & 0.234\% \\
\hline
\hline
integrating multiple density & 65.2\% & 90.0\% & 96.1\% &- & 0.553\% & 0.403\% & 0.197\% \\
\hline
\end{tabular}
\label{tab:errors}
\end{table*}

\begin{figure}[ht]
  \centering
  \includegraphics[width=0.70\linewidth]{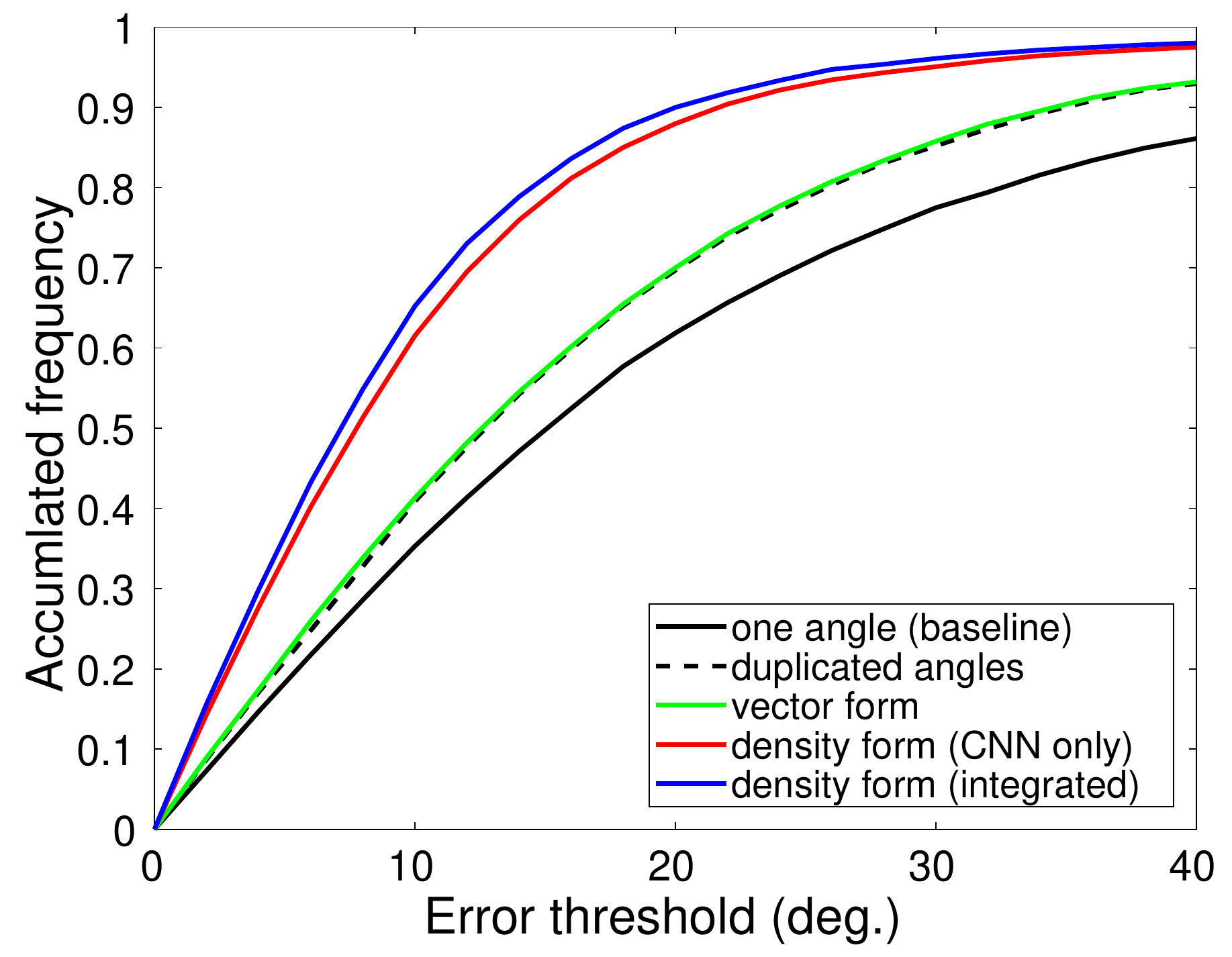}  
  \caption{Visual comparison of the orientation error populations.}
  \label{fig:errors}
\end{figure}
  
\begin{figure}[t]
  \centering
  \begin{tabular}{cc}
    \includegraphics[width=0.35\linewidth,bb=0 0 240 300 ]{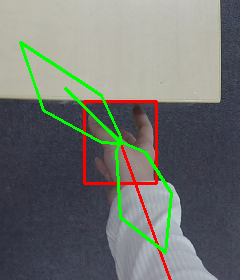} &
    \includegraphics[width=0.35\linewidth,bb=0 0 240 300 ]{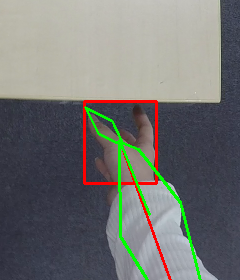} \\ 
  \end{tabular}
  \caption{Improvement by integration. (left) large orientation error given by $p(\theta_i|H_{img})$. (right) correct orientation given by $p(\theta_i|H_{img},H_r,cx,cy)=p(\theta_i|H_{img})p(\theta_i|H_r)p(\theta_i|cx,cy)$.}
  \label{fig:incorrect_results}
\end{figure}

\subsection{Results and discussion}
\figrefh{fig:correct_results} illustrates the estimated results on the group work FPV dataset when applying probability density representation $p(\theta_i|H_{img})$ to orientation estimation  that uses only the hand region images. We can see that the radar chart visualization of the probability density is strongly biased to the wrist direction, and the final orientation results indicated by the green lines are close to the ground truth indicated by the red lines.

For statistical analysis, we calculated the percentages of samples whose estimation errors were within particular angles. 
The performance comparison in \tabref{tab:errors} and \figref{fig:errors} shows those percentage values among the validated three methods.
Even the proposed method using only hand region images achieved a significant improvement in the ``less than $10^\circ, 20^\circ, 30^\circ$'' categories compared with the conventional methods.
We have two hypothetic reasons for the accuracy improvement.
\begin{enumerate}
\item The conventional methods try to find correct orientation, but not to omit incorrect orientation. In contrast, the probablity density approach achieves both both them, because CNN training proceeds to increase probablity values at $\theta_i$ close to the ground truth and to suppress probablity values at other orientations, simultaneously.
\item  Orientation is not a label, but a scalar. Additionally, nobady can know (or define) pricise hand orientation values under various hand pose including finger form and connection to wrist joint. Ground truth values have certain ambiguity due to those reasons. The probability distribution given by Eq. (1) successfully handles this issue. Note that correct $\sigma$ should be configured to get this effect.
\end{enumerate}
  
%Using the vector representation method, the estimated error directly affected the orientation error, whereas using the probability density method, the final prediction was determined by a relative comparison of likelihoods among multiple orientations.
%Therefore, the estimation errors in the probability density sometimes did not have a significant influence on the final prediction, for example, when similar values of estimation errors were caused in multiple orientations or when small errors were caused at distant orientations from the ground truth.

However, the proposed method slightly increased the frequencies in the ``more than $90^\circ,120^\circ,150^\circ$'' categories.
This disadvantage was also caused by the use of the probability density form. Essentially, the $N$-path CNN individually predicted $N$ values, even if the softmax normalization at its last layer  provided some dependency among them.  This characteristic may deliver multiple peaks at distant orientations in $p(\theta_i)$, despite being inconsistent with the actual scenario. 
As an example of an extreme case for intuitive understanding, assume two similar hand images with quite different ground truth orientations are provided for training.
In the probability density scheme, the CNN parameters are optimized to output two distinct peaks at the two ground truth orientations. When an incorrect peak is selected as a final prediction, its error becomes large. 
By contrast, the CNN that uses the vector representation is trained to output the mean of the two ground truth orientations, which can avoid large orientation errors.
This analysis is supported by the typical failure example shown on the left of \figref{fig:incorrect_results}.  The estimated probability density has two peaks at distant orientations in the estimated probability density: one is close to the ground truth, but the other is quite far from it.

The above disadvantage can be  reduced by integrating multiple estimations, as shown on the right of \figref{fig:incorrect_results}. Particularly, the human region-based method constrains the possible range of hand orientation, which suppresses incorrect selection from multiple peaks in $p(\theta_i|H_{img})$. The integration also provides a further improvement in the ``less than $10^\circ, 20^\circ, 30^\circ$'' categories, as shown in \tabref{tab:errors}. The proposed scheme provides a choice of integration for further accuracy.
However, the additional two methods also have issues. The human region does not work well when the hand lies in the body region. Additionally, we confirmed that the hand orientation prior based on its position strongly depends on the scene, such as discussion using a white board and cooperative assembly work. The simple product of the probability densities assumes that all the integration targets have the same confidence and does not consider the weights among them.

%--------------------------------------------------------------
\section{Conclusion}
In this paper, we proposed predicting hand orientation in probability density form. On the group work FPV dataset, the proposed method estimated hand orientation more accurately than the conventional vector representation method. Additionally, we easily integrated multiple results represented in the probability density form and confirmed that it leads to improved performance.
For further accuracy, the probability density form should be applied to an advanced CNN structure with a fusion of multiple scale features. The additional estimation methods, including their integration scheme, must be improved.
%In the near future, we will analyze interactions via hand behaviors in group work and provide meaningful feedback to the education technology field.

{\small 
\bibliographystyle{ieee_fullname}
%\bibliography{ref} % for nomal use

     % for arxiv ( run bibtex to generate .bbl and use it like this. ref.bib is unnecessary 
}

\end{document}